\documentclass{article}
\usepackage[margin=1in]{geometry}  
\usepackage{graphicx}
\usepackage{booktabs}
\usepackage[most]{tcolorbox}
\usepackage{url}
\usepackage{subcaption}

\title{\vspace{-1cm} 
Lecture Video Visual Objects (LVVO) Dataset: \\ 
A Benchmark for Visual Object Detection in Educational Videos \\ \vspace{0.0cm}}

\author{
Dipayan Biswas \hspace{0.5cm} Shishir Shah \hspace{0.5cm} Jaspal Subhlok \\
Department of Computer Science, University of Houston, Houston, TX \\
\vspace{0.5em}
{\small \url{https://github.com/dipayan1109033/LVVO_dataset}}
}

\date{}

\begin{document}

\maketitle

\begin{abstract}
We introduce the Lecture Video Visual Objects (LVVO) dataset, a new benchmark for visual object detection in educational video content. The dataset consists of 4,000 frames extracted from 245 lecture videos spanning biology, computer science, and geosciences. A subset of 1,000 frames, referred to as \texttt{LVVO\_1k}, has been manually annotated with bounding boxes for four visual categories: \textit{Table}, \textit{Chart-Graph}, \textit{Photographic-image}, and \textit{Visual-illustration}. Each frame was labeled independently by two annotators, resulting in an inter-annotator F1 score of 83.41\%, indicating strong agreement. To ensure high-quality consensus annotations, a third expert reviewed and resolved all cases of disagreement through a conflict resolution process. To expand the dataset, a semi-supervised approach was employed to automatically annotate the remaining 3,000 frames, forming \texttt{LVVO\_3k}. The complete dataset offers a valuable resource for developing and evaluating both supervised and semi-supervised methods for visual content detection in educational videos. The LVVO dataset is publicly available to support further research in this domain.
\end{abstract}

\section{Introduction}
The Lecture Video Visual Objects (LVVO) Dataset is designed as a benchmark for object detection in lecture video frames. It includes bounding box annotations for four visual categories: Table, Chart-Graph, Photographic-Image, and Visual-Illustration.

The dataset consists of 4,000 images (video frames) extracted from a diverse collection of lecture videos. Out of these, a randomly selected subset of 1,000 images has been manually annotated by expert annotators, forming the \texttt{LVVO\_1k} labeled dataset. Each image was independently annotated by two annotators, with a third expert reviewing and resolving any disagreements to ensure high-quality consensus annotations. The following sections detail the dataset creation process and present key statistics gathered during its development.

\section{Dataset Preparation}
To build our dataset, we collected lecture videos from \textit{videopoints.org} \cite{videopoints}. We then extracted 4,000 visually rich and distinct frames, ensuring diversity across multiple instructors and subject areas.

\subsection{Lecture Video Collection}
The lecture videos were sourced from \textit{videopoints.org} \cite{videopoints}, a platform hosting screen-captured live lectures, as part of the previous work in \cite{rahman2023enhancing}. The collection includes videos from eight different instructors, covering 13 distinct courses, with a total of 245 lecture videos. These lectures span three subject areas: biology, computer science, and geosciences. To ensure the inclusion of the most recent lectures, we selected courses from the latest semesters offered by each instructor. The lectures in the dataset were recorded between 2019 and 2024.

\subsection{Unique Frame Extraction}
We adopted the method from \cite{tuna:learningCompanion} to identify slide transition points and extract key frames representing distinct slides from the lecture videos. However, we observed duplicate frames, often caused by instructors revisiting previous slides during lectures. To address this, we extended the algorithm to detect and remove duplicate frames within a window of key frames for each video. Additionally, we prioritized filtering out frames that contained only textual content with no significant visual elements. These refinements ensured that the final dataset retained unique video frames with significant visual content, resulting in a finalized set of 4,000 images. Each image file is named using the format: 
\texttt{<instructor\_id>\_<course\_id>\_<video\_id>\_<filename>}.
Table~\ref{tab:frame_statistics} summarizes the distribution of instructors, courses, and extracted frames across the three subject areas.

\begin{table}[htb]
    \centering
    \caption{Distribution of Instructors, Courses, and Extracted Video Frames Across Subject Areas}
    \resizebox{\textwidth}{!}{%
    \begin{tabular}{@{}lcccc@{}}
        \toprule
        \textbf{Subject} & \textbf{\# Instructors} & \textbf{\# Courses} & \textbf{Total Frames Count} & \textbf{Manually Annotated (LVVO\_1k)} \\ 
        \midrule
        \textbf{Biology}          & 3 & 6 & 1,932& 408 \\
        \textbf{Computer Science} & 3 & 4 & 786 & 245 \\
        \textbf{Geosciences}      & 2 & 3 & 1,282 & 347 \\ 
        \midrule
        \textbf{Total}            & 8 & 13 & 4,000 & 1,000 \\ 
        \bottomrule
    \end{tabular}%
    }
    \label{tab:frame_statistics}
\end{table}

\section{Manual Annotation}
\subsection{Annotation Workflow}
A randomly selected subset of 1,000 images (referred to as \texttt{LVVO\_1k}) was manually labeled by expert annotators with bounding box annotations for four distinct categories: Table, Chart-Graph, Photographic-Image, and Visual-Illustration. Annotating lecture slides presents unique challenges. They typically consist of artificially designed visual content where visual objects have diverse semantic meanings and weak structural boundaries—unlike well-defined objects in natural images such as chairs, tables, cats, or dogs \cite{biswas2023identifyobject}. To ensure high-quality and consistent annotations, we engaged graduate students with relevant domain expertise and provided them with unified instructions (see Section~\ref{annotation-instructions}). The annotation was carried out using the Microsoft VoTT annotation tool \cite{VoTT}, which allowed annotators to draw bounding boxes and assign category labels. The process followed three phases:

\begin{enumerate}
    \item \textbf{Initial Calibration:} All the annotators labeled an initial set of 50 sample frames using the provided instructions. 
    After annotation, a group discussion was conducted to review differences and understand challenges in annotation. Subsequently, the guidelines were modified to reduce ambiguity. 
    \item \textbf{Independent Annotation:} The remaining frames were divided among the annotators, with each frame independently labeled by two annotators, following the finalized guidelines to ensure cross-verification and consistency.
    \item \textbf{Conflict Resolution:} A third expert was involved only in cases where the initial two annotators disagreed. For each such instance, the expert resolved conflicts by selecting the most accurate bounding boxes from one or both annotators, thereby finalizing the annotation.

\end{enumerate}

This rigorous annotation process ensures the dataset’s reliability and consistency, making it well-suited for benchmarking object detection models on lecture video frames.

\subsection{Annotation Instructions}
\label{annotation-instructions}
The following instructions were provided to the annotators to ensure consistency and accuracy during the manual labeling process. These guidelines define what qualifies as a visual object, outline the annotation procedure, and specify the categories used for labeling. Annotators followed these instructions while using the VoTT annotation tool \cite{VoTT} to perform bounding box annotations on the selected video frames.

\begin{tcolorbox}[colback=gray!5, colframe=gray!50, breakable]

\textbf{Task:} To identify and categorize visual objects in video frames that are meaningful to the video content. Specifically, you will:

\begin{enumerate}
    \item Identify and draw a bounding box around each visual object.
    \item Label each identified visual object with a category selected from the provided list below.
\end{enumerate}

\textbf{What is a Visual Object?}  
For the purpose of this task, a visual object contains an image or multiple images that together represent \textit{meaningful semantic content} in the video.

\begin{itemize}
    \item Visual objects can be photographic images, charts, tables, or illustrations. A visual object may contain text such as the content of the cells in a table or labels of components in the image. It should not include captions or descriptions that are not directly a part of the image.
    \item Images that are not relevant to the lecture content are not considered visual objects. For example, speaker faces, logos, and other content that is part of the video frame background should not be selected as visual objects.
    \item Your goal is to select \textit{coherent and complete} visual objects, that we refer as \texttt{valid} objects. In some cases, a larger visual object consisting of nearby \texttt{valid} visual objects may also appear to be a \texttt{valid} visual object. In such situations, it is sufficient to select only the smaller \texttt{valid} visual objects.
    \item The rectangular bounding boxes may overlap, but the visual objects themselves should not.
\end{itemize}

\textbf{Categories:} Assign one of the following category labels to each visual object you identify:

\begin{itemize}
    \item \textbf{Table:} An arrangement of information or data, typically in rows and columns.
    \item \textbf{Chart-Graph:} Graphical representation of data.
    \item \textbf{Photographic-Image:} Pictures that are made using cameras.
    \item \textbf{Visual-Illustration:} Diagrams, flowcharts, and other visual illustrations.
\end{itemize}

In following these steps, use your best judgment in case of ambiguity. In some cases, the boundaries, the category label, or even the existence of a visual object may not be clear. We are looking forward to your best guess in such scenarios.

\end{tcolorbox}

\subsection{Annotation Statistics}

\subsubsection{Comparison of Independent Annotations}
\begin{figure}[tb!]
    \centering
    \includegraphics[width=\linewidth]{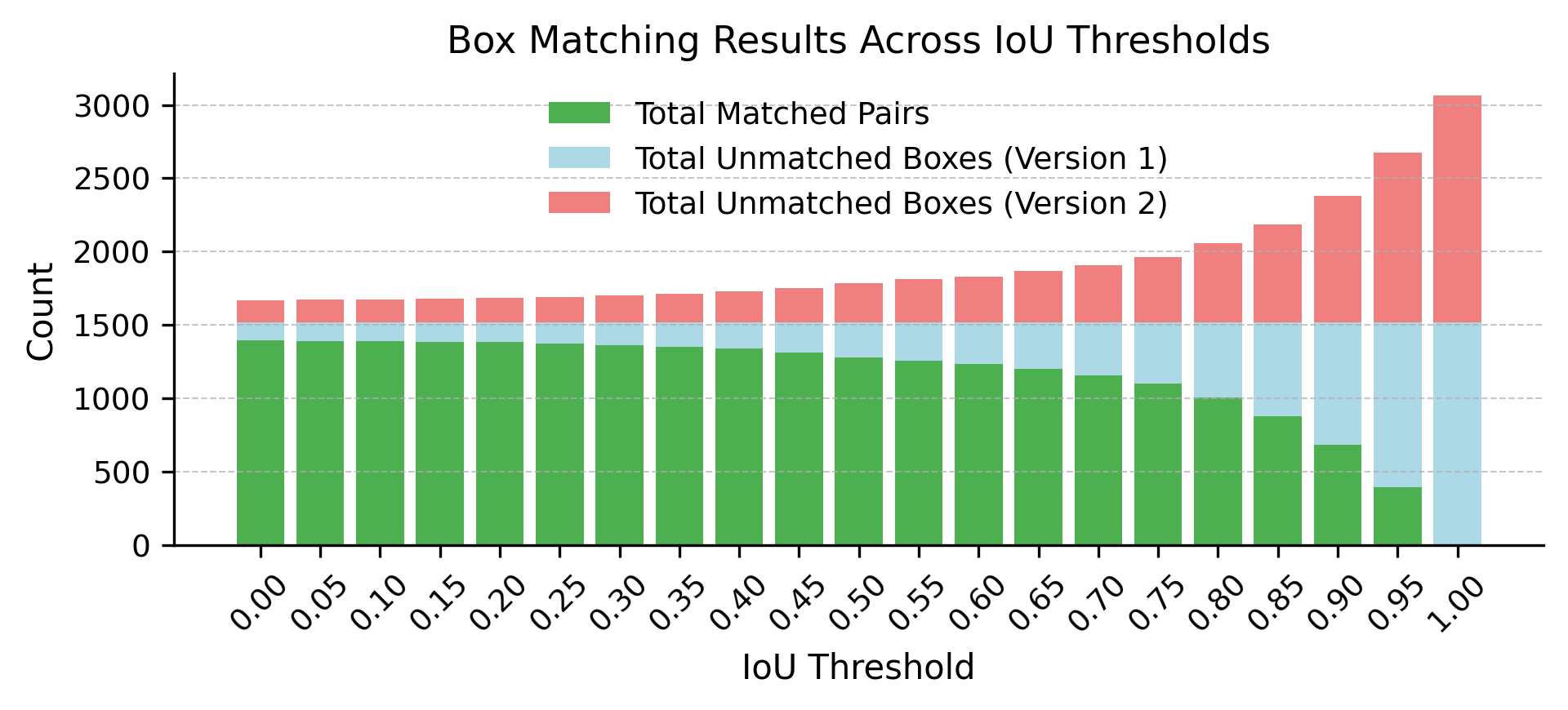}
    \caption{Stacked bar plot showing matched (green) and unmatched boxes (blue: version 1, red: version 2) over IoU thresholds. Higher thresholds lead to fewer matches due to stricter overlap requirements.}
    \label{fig:iou_stacked_bar}
\end{figure}

To assess the agreement of independent annotations, we compared the two versions in which each image was labeled by different annotator. For each frame, bounding boxes from the two annotation sets were matched using a greedy algorithm that iteratively selects box pairs with the highest Intersection over Union (IoU), ensuring that each box is matched only once. The process continues until no remaining pairs meet the provided IoU threshold. After completing the matching across all frames, we aggregated the total number of matched pairs and unmatched boxes for each version. Figure~\ref{fig:iou_stacked_bar} presents a stacked bar chart showing the distribution of matched and unmatched boxes across a range of IoU thresholds. 

Each bar corresponds to a specific IoU threshold value, with the total height representing the combined count of matched pairs and unmatched boxes. The green segment indicates the number of matched pairs, while the blue and red segments represent unmatched boxes from version 1 and version 2, respectively. 
Moving from right to left (i.e., from high to low IoU thresholds), there is a significant increase in the number of matched pairs at higher IoU values. This indicates that most matches occur when bounding boxes are closely aligned—suggesting that the two annotators generally placed bounding boxes in similar positions. As the IoU threshold decreases, the number of matched pairs gradually declines and eventually levels off. At the lowest thresholds, there remain some unmatched boxes—specifically, 122 from version 1 and 152 from version 2. These likely reflect semantic disagreements or differing interpretations between annotators.
At an IoU threshold of 0.5, 1278 matched pairs (involving 2556 boxes) were identified, with 239 and 269 unmatched boxes from versions 1 and 2, respectively—reflecting an 83.41\% agreement and strong annotator alignment under moderate overlap conditions.

\begin{figure}[tb!]
    \centering
    \includegraphics[width=0.60\linewidth]{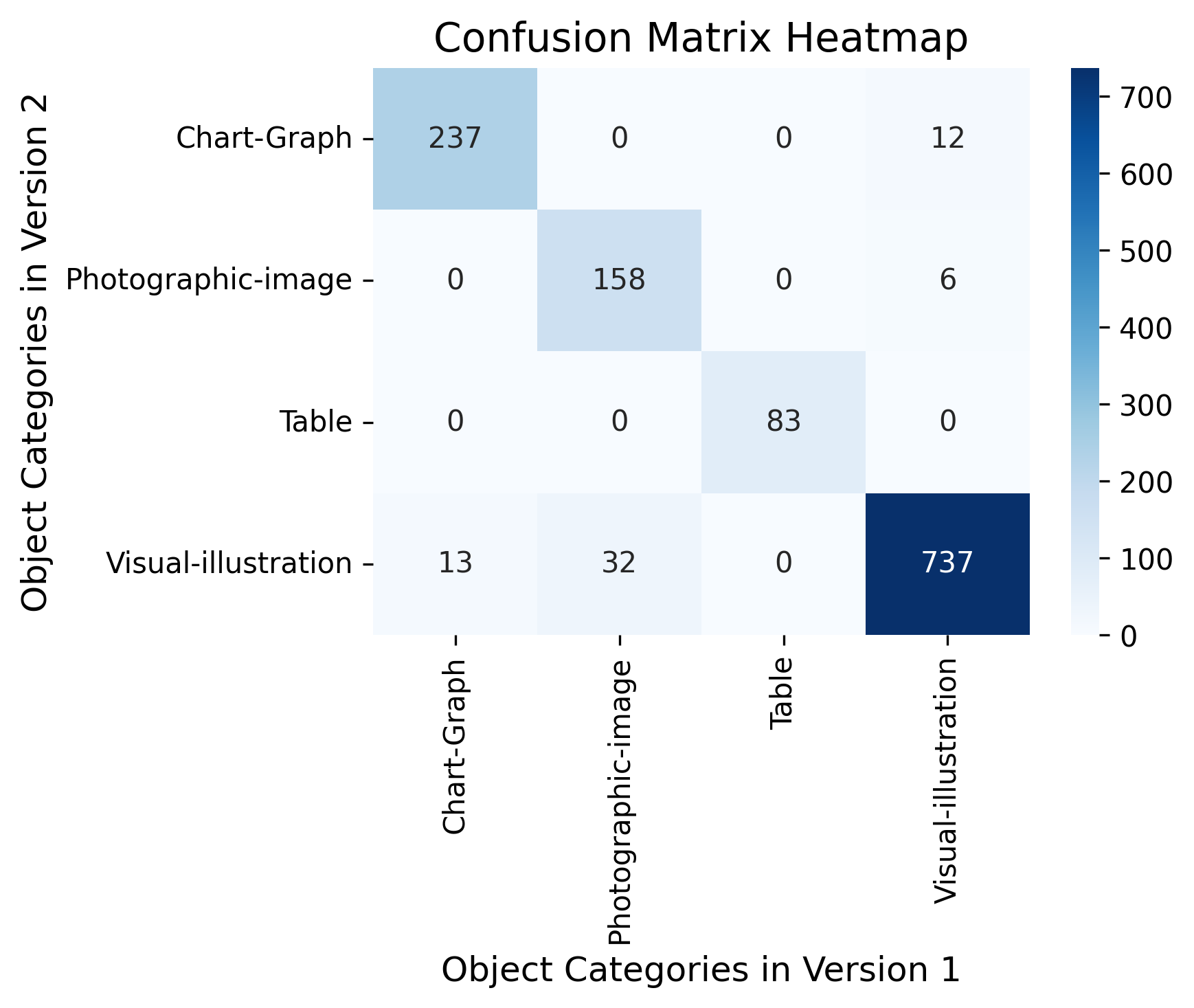}
    \caption{Confusion matrix showing category-wise agreement among matched annotations (where IoU $>=$ 0.75). Strong diagonal values indicate high consistency, while off-diagonal elements reveal label discrepancies.}
    \label{fig:confusion_matrix_versions}
\end{figure}

Figure \ref{fig:confusion_matrix_versions} illustrates the category-wise agreement between version 1 and version 2 annotations, limited to matched pairs with an IoU threshold of at least 0.75. The strong diagonal dominance indicates high labeling consistency across versions, while the sparse off-diagonal entries highlight rare instances of category mismatch. Notably, the \textit{Visual-illustration} category is the most commonly confused with \textit{Chart-Graph} and \textit{Photographic-image}, suggesting occasional ambiguity in distinguishing between these categories.

Although the annotations showed strong overall agreement, discrepancies were still present. To address these discrepancies, a third expert was involved to review and resolve conflicts, as outlined before.

\subsubsection{Final Annotation}
Following the conflict resolution process, we finalized the consensus annotations used in the dataset.

\begin{figure*}[!ht]
  \centering
  \begin{subfigure}[b]{0.45\textwidth}
      \includegraphics[width=\linewidth]{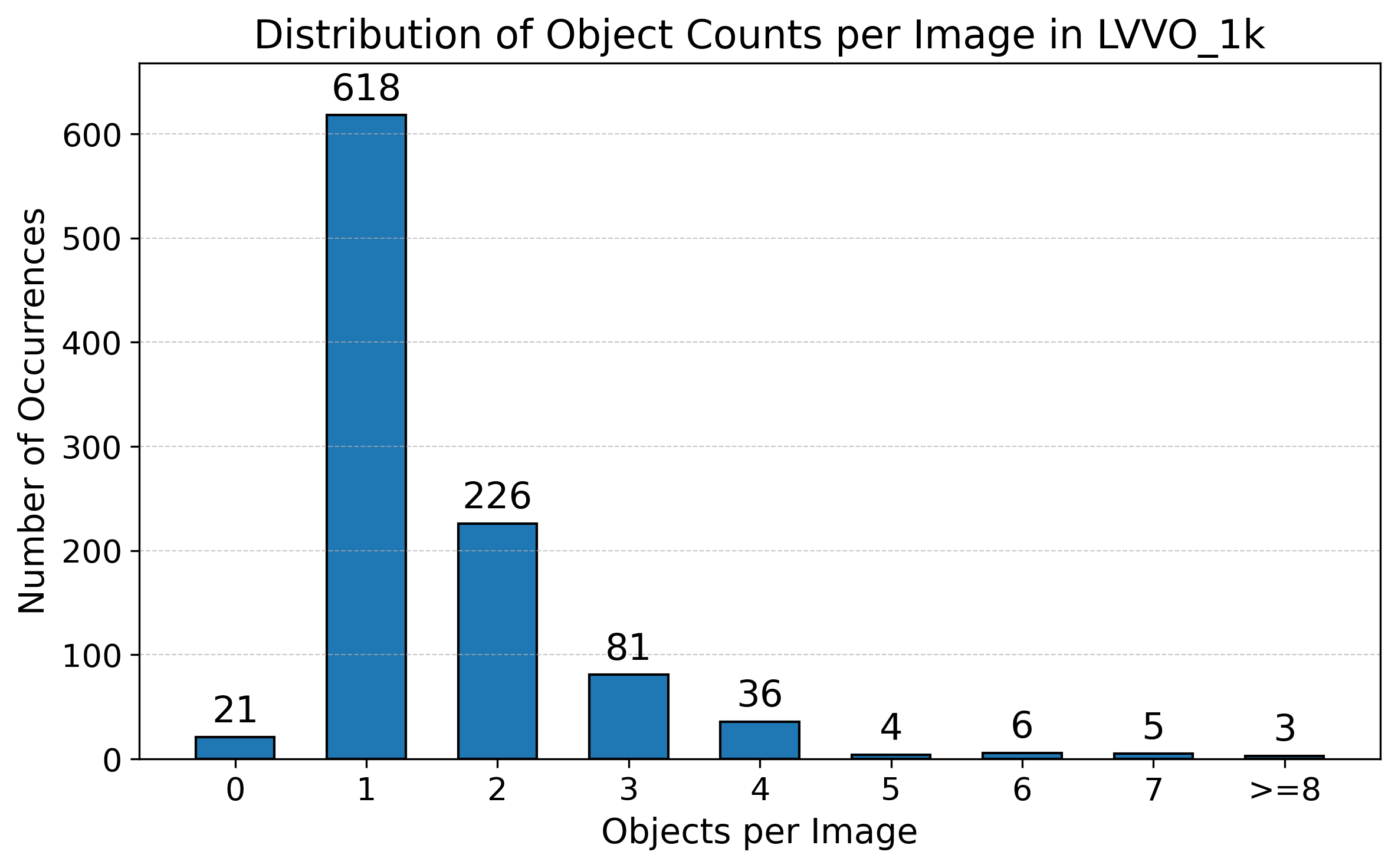}
      \caption{Object counts per image.}
      \label{fig:consensus-objects}
  \end{subfigure}
  \hfil
  \begin{subfigure}[b]{0.45\textwidth}
      \includegraphics[width=\linewidth]{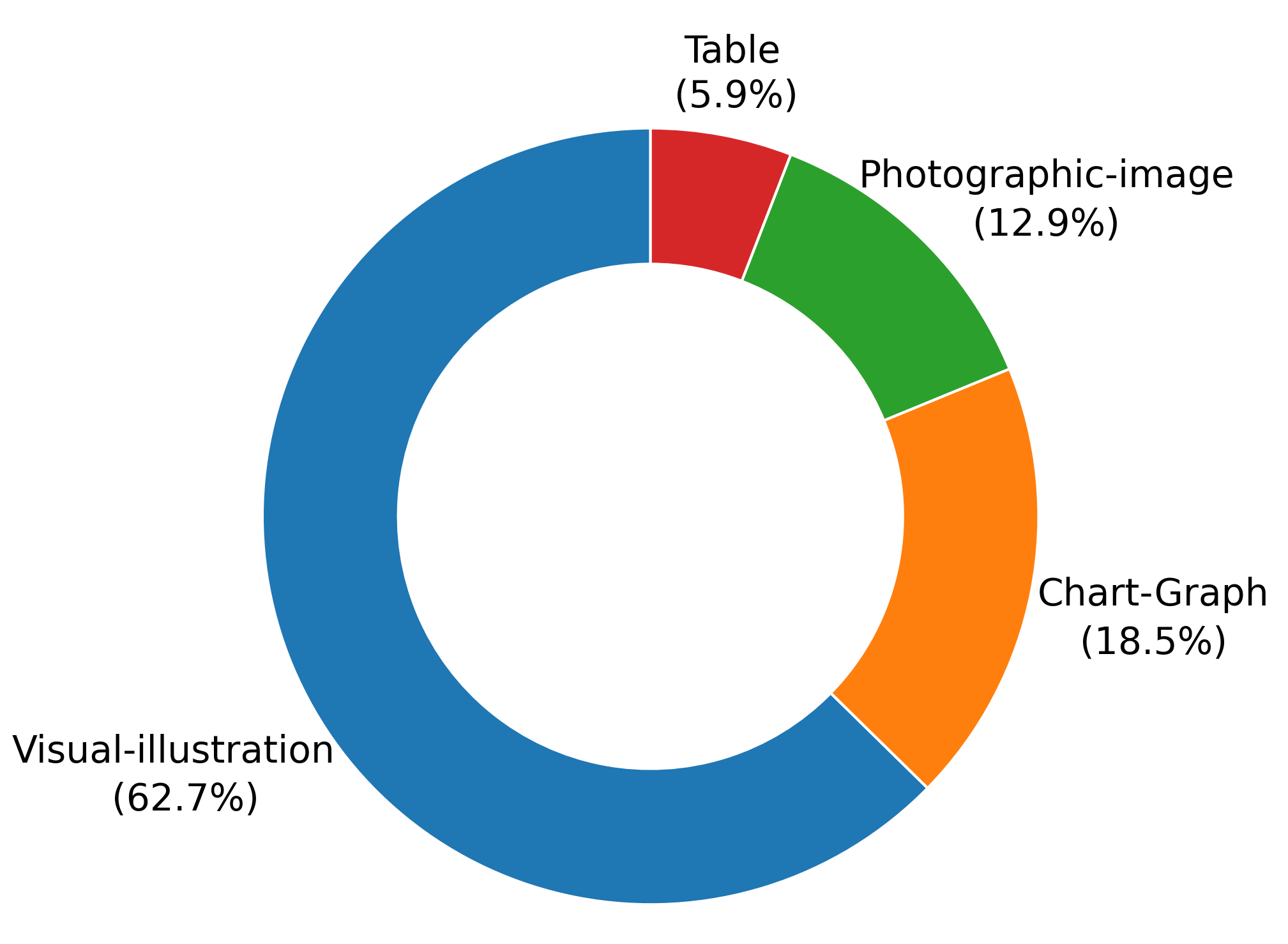}
      \caption{Distribution of object categories.}
      \label{fig:consensus-classes}
  \end{subfigure}
  
  \caption{Summary of consensus annotations after conflict resolution. (a) Distribution of the number of annotated objects per image. (b) Category-wise distribution of annotated objects.}
  \label{fig:consensus_stats}
\end{figure*}

Figure~\ref{fig:consensus-objects} shows that most images contain one or two objects, with fewer images containing higher counts. Figure~\ref{fig:consensus-classes} highlights that Visual-illustration dominates the category distribution, followed by Chart-Graph and Photographic-image, with Table being the least common.

\section{Automatic Annotation}
The LVVO dataset contains 4,000 video frames, of which only 1,000 were manually labeled because of the significant effort required for high-quality manual annotation. To expand the \textit{LVVO} dataset further and reduce manual effort, we are also releasing the remaining 3,000 frames, which have been automatically labeled using the methodology described below.

\vspace{0.1cm} 

Our approach \cite{biswas2025visual} involves fine-tuning a COCO-pretrained YOLOv11 \cite{yolo11_ultralytics} model using transfer learning. The model is first adapted to the manually annotated \texttt{LVVO\_1k} dataset, with an 80\% training and 20\% validation split. Once fine-tuned, the model is used in inference mode to predict bounding boxes on the unlabeled images. A confidence threshold of 0.5 is applied to discard low-confidence predictions, ensuring the quality of the automatically generated annotations. This automatic labeling process results in the \texttt{LVVO\_3k} automatically labeled dataset. Combined with the manually annotated portion, it expands the \textit{LVVO\_dataset} to a total of 4,000 labeled images, supporting further model development and evaluation.

\section{Dataset Files Description}

Here, the files and structure associated with the \texttt{LVVO} dataset are described. Three dataset variants are provided, each following a consistent internal structure:

\begin{itemize}
    \item \textbf{\texttt{LVVO\_1k\_withCategories.zip}}: The manually annotated subset containing 1,000 images with the associated category labels.
    \item \textbf{\texttt{LVVO\_1k.zip}}: The same 1,000-image subset as above, but with all objects treated as a single category (a generic class label: \texttt{object}).
    \item \textbf{\texttt{LVVO\_3k.zip}}: The automatically annotated subset containing 3,000 additional images.
\end{itemize}

Each dataset archive includes the following components:

\begin{itemize}
    \item \textbf{\texttt{images/}}: Contains the image files.
    \item \textbf{\texttt{labels/}}: Contains JSON annotation files. Each file shares the same base name as its corresponding image file, allowing one-to-one mapping.
    \item \textbf{\texttt{dataset\_info.json}}: Contains metadata including category names and their corresponding IDs, as well as mappings between image filenames and unique image identifiers.
\end{itemize}


\end{document}